\documentclass[10pt,twocolumn,letterpaper]{article}

\usepackage{iccv}
\usepackage{times}
\usepackage{epsfig}
\usepackage{graphicx}
\usepackage{array}
\usepackage{amsmath}
\usepackage{amssymb}
\usepackage{color, xcolor,colortbl}
\usepackage{booktabs}
\usepackage{makecell}
\usepackage{multirow}
\usepackage[square,comma,numbers,sort&compress]{natbib}
\usepackage[normalem]{ulem}

\newcommand{\vv}[1]{{\texttt{#1}}}
\newcommand{\conv}{\vv{conv}}
\newcommand{\fc}{\vv{fc}}
\newcolumntype{x}{>\small c}

\newcommand{\trainval}{trainval\raisebox{0.2ex}{$\ast$}}
\newcommand{\minival}{minival\raisebox{0.2ex}{$\ast$}}
\newcolumntype{L}[1]{>{\raggedright\let\newline\\\arraybackslash\hspace{0pt}}m{#1}}
\newcolumntype{C}[1]{>{\centering\let\newline\\\arraybackslash\hspace{0pt}}m{#1}}
\newcolumntype{R}[1]{>{\raggedleft\let\newline\\\arraybackslash\hspace{0pt}}m{#1}}
\newcommand{\hl}[1]{\underline{\textbf{#1}}}
\newcolumntype{o}{>\small L}

\makeatletter
\def\maxwidth#1{\ifdim\Gin@nat@width>#1 #1\else\Gin@nat@width\fi}
\makeatother

\usepackage[pagebackref=true,breaklinks=true,letterpaper=true,colorlinks,bookmarks=false]{hyperref}

\iccvfinalcopy 

\def\iccvPaperID{2587} 

\ificcvfinal\fi
\begin{document}

\title{Beyond Skip Connections: Top-Down Modulation for Object Detection}

\author{}
\author{Abhinav Shrivastava$^{1,3}$ \quad\quad Rahul Sukthankar$^3$ \quad\quad Jitendra Malik$^{2,3}$ \quad\quad Abhinav Gupta$^{1,3}$\\[0.1in]
$^1$Carnegie Mellon University \quad
$^2$University of California, Berkeley \quad
$^3$Google Research
\\}

\maketitle
\begin{abstract}
In recent years, we have seen tremendous progress in the field of object detection. Most of the recent improvements have been achieved by targeting deeper feedforward networks. However, many hard object categories such as bottle, remote, etc.\ require representation of fine details and not just coarse, semantic representations. But most of these fine details are lost in the early convolutional layers. What we need is a way to incorporate finer details from lower layers into the detection architecture. Skip connections have been proposed to combine high-level and low-level features, but we argue that selecting the right features from low-level requires top-down contextual information. Inspired by the human visual pathway, in this paper we propose top-down modulations as a way to incorporate fine details into the detection framework. Our approach supplements the standard bottom-up, feedforward ConvNet with a top-down modulation (TDM) network, connected using lateral connections. 
These connections are responsible for the modulation of lower layer filters, and the top-down network handles the selection and integration of contextual information and low-level features. The proposed TDM architecture provides a significant boost on the COCO benchmark, achieving 28.6 AP for VGG16 and 35.2 AP for ResNet101 networks. 
Using InceptionResNetv2, our TDM model achieves \textbf{37.3} AP, which is the best single-model performance to-date on the COCO testdev benchmark, without any bells and whistles.
\vspace{-0.1in}
\end{abstract}

\section{Introduction}\label{sec:intro}
\begin{figure}
    \centering
    \includegraphics[width=\linewidth]{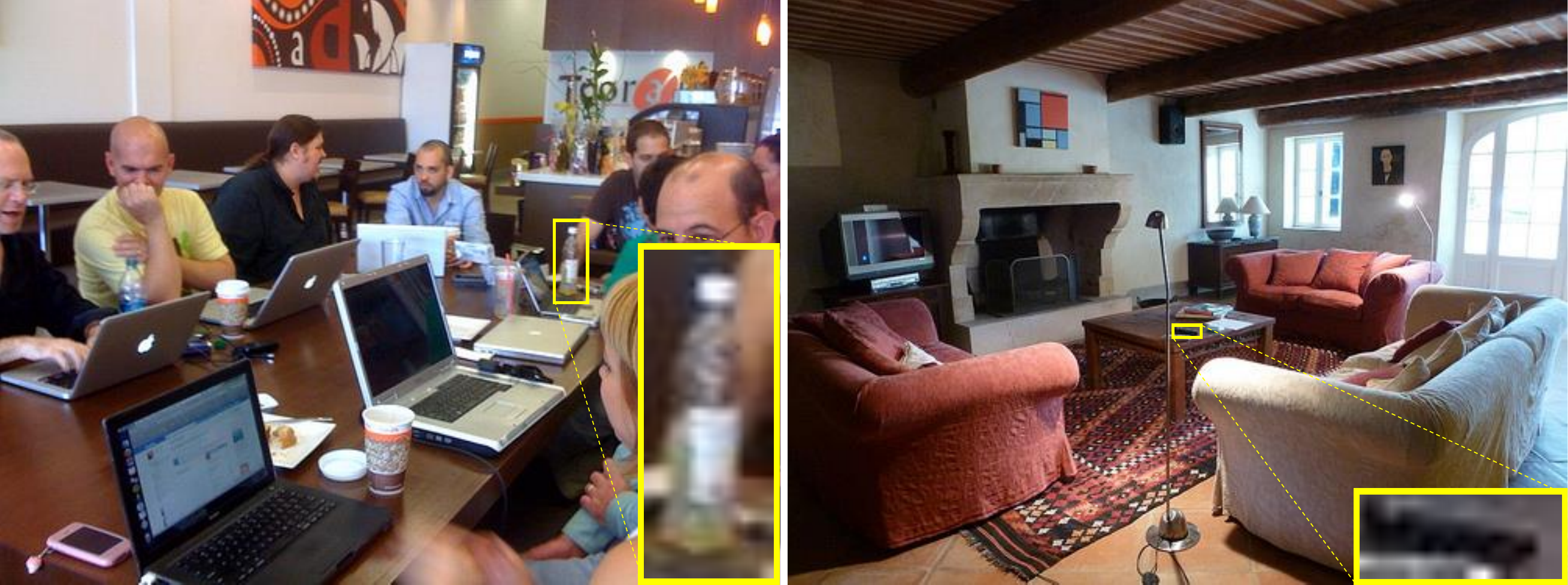}
    \caption{Detecting objects such as the bottle or remote shown above requires low-level finer details as well as high-level contextual information. In this paper, we propose a top-down modulation (TDM) network, which can be used with any bottom-up, feedforward ConvNet. We show that the features learnt by our approach lead to significantly improved object detection.}
    \label{fig:bottleRemote}
    \vspace{-0.2in}
\end{figure}

Convolutional neural networks (ConvNets) have revolutionized the field of object detection~\cite{rcnn,girshick2015fast,ren2015faster,szegedy2013deep,gidaris2015object,shrivastavaOHEM,bell2015inside}. Most standard ConvNets are bottom-up, feedforward architectures constructed using repeated convolutional layers (with non-linearities) and pooling operations~\cite{alexnet,VGG,resnet,googlenet,szegedy2016inception}. These convolutional layers learn invariances and the spatial pooling increases the receptive field of subsequent layers; thus resulting in a coarse, highly semantic representation at the final layer.

However, consider the images shown in Figure~\ref{fig:bottleRemote}. Detecting and recognizing an object like the bottle in the left image or remote in the right image requires extraction of very fine details such as the vertical and horizontal parallel edges. But these are exactly the type of edges ConvNets try to gain invariance against in early convolutional layers. One can argue that ConvNets can learn not to ignore such edges when in context of other objects like table. However, objects such as table do not emerge until very late in feed-forward architecture. So, how can we incorporate these fine details in object detection?

A popular solution is to use variants of `skip' connections~\cite{sermanet2013pedestrian,farabet2013learning,bell2015inside,hariharan2015hypercolumns,long2015fully,xie2015holistically}, that capture these finer details from lower convolutional layers with \emph{local} receptive fields. But simply incorporating high-dimensional skip features into detection does not yield significant improvements due to overfitting caused by curse of dimensionality. What we need is a selection/attention mechanism that selects the relevant features from lower convolutional layers.

We believe the answer lies in the process of \emph{top-down modulation}. In the human visual pathway, once receptive field properties are tuned using feedforward processing, \emph{top-down modulations} are evoked by feedback and horizontal connections~\cite{lamme1998feedforward,kravitz2013ventral}. These connections modulate representations at multiple levels~\cite{zanto2010top,zanto2011causal,gazzaley2012top,piech2013network,gilbert2007brain} and are responsible for their selective combination~\cite{hopfinger2000neural,chun1999top}. We argue that the use of skip connections is a special case of this process, where the \emph{modulation} is relegated to the final classifier, which directly tries to influence lower layer features and/or learn how to combine them.

In this paper, we propose to incorporate the \emph{top-down modulation} process in the ConvNet itself. Our approach supplements the standard bottom-up, feedforward ConvNet with a top-down network, connected using lateral connections. These connections are responsible for the \emph{modulation} and \emph{selection} of the lower layer filters, and the top-down network handles the \emph{integration} of features.

Specifically, after a bottom-up ConvNet pass, the final high-level semantic features are transmitted back by the top-down network. Bottom-up features at intermediate depths, after lateral processing, are combined with 
the top-down features, and this combination is further transmitted down by the top-down network. Capacity of the new representation is determined by lateral and top-down connections, and optionally, the top-down connections can increase the spatial resolution of features. These final, possibly high-res, top-down features inherently have a combination of \emph{local} and \emph{larger} receptive fields. 

The proposed Top-Down Modulation (TDM) network is trained end-to-end and can be readily applied to any base ConvNet architecture (\eg, VGG~\cite{VGG}, ResNet~\cite{resnet}, Inception-Resnet~\cite{szegedy2016inception} \etc). To demonstrate its effectiveness, we use the proposed network in the standard Faster R-CNN detection method~\cite{ren2015faster} and evaluate on the challenging COCO benchmark~\cite{lin2014coco}. We report a consistent and significant boost in performance on all metrics across network architectures. 
TDM network increases the performance of vanilla Faster R-CNN with: (a) VGG16 from 23.3 AP to textbf{28.6} AP, (b) ResNet101 from 31.5 AP to \textbf{35.2} AP, and (c) InceptionResNetv2 from 34.7 AP to \textbf{37.2} AP. These are the best performances reported to-date for these architectures without any bells and whistles (\eg, multi-scale features, iterative box-refinement).
Furthermore, we see drastic improvements in small objects (\eg, $+$\textbf{4.5} AP) and in objects where selection of fine details using top-down context is important.

\begin{figure*}
    \centering
        \vspace{-0.1in}
    \includegraphics[width=0.98\linewidth]{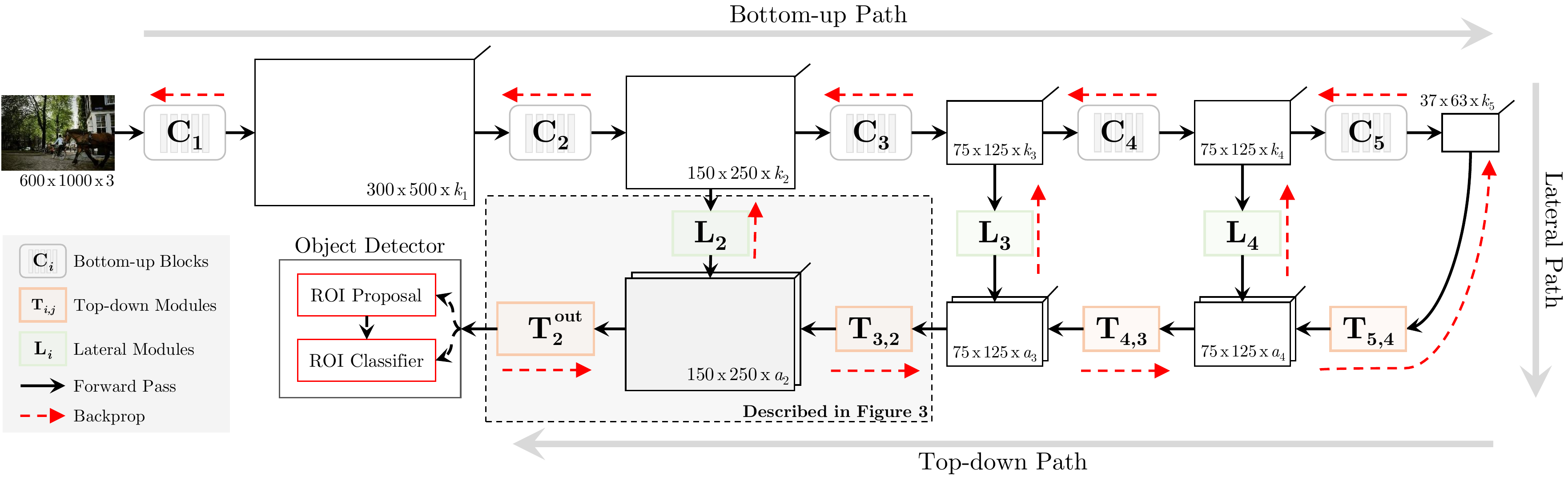}
            \vspace{-0.07in}
    \caption{The illustration shows an example of \textbf{Top-Down Modulation} (TDM) Network, which is integrated with the bottom-up network with lateral connections. $\mathbf{C}_i$ are bottom-up, feedforward feature blocks, $\mathbf{L}_i$ are the lateral modules which transform low level features for the top-down contextual pathway. Finally, $\mathbf{T}_{j,i}$, which represent flow of top-down information from index $j$ to $i$. Individual components are explained in Figure~\ref{fig:overview_short} and~\ref{fig:overview_details}.
    }
    \vspace{-0.2in}
    \label{fig:overview_long}
\end{figure*}

\vspace{-0.01in}
\section{Related Work}\label{sec:related}
\vspace{-0.01in}

After the resurgence of ConvNets for image classification~\cite{alexnet,imagenet}, they have been successfully adopted for a variety of computer vision tasks such as object detection~\cite{rcnn,girshick2015fast,ren2015faster,szegedy2013deep}, semantic segmentation~\cite{deeplab,liu2015parsenet,long2015fully,badrinarayanan2015segnet}, instance segmentation~\cite{sds,hariharan2015hypercolumns,pinheiro2015learning}, pose estimation~\cite{toshev2014deeppose,tompson2014joint}, depth estimation~\cite{eigen2015predicting,wang2015designing}, edge detection~\cite{xie2015holistically}, optical flow predictions~\cite{fischer2015flownet,ranjan2016optical} \etc. 
However, by construction, final ConvNet features lack the finer details that are captured by lower convolutional layers. These finer details are considered necessary for a variety of recognition tasks, such as accurate object localization and segmentation.

To counter this, `skip' connections have been widely used with ConvNets. Though the specifics of methods vary widely, the underlying principle is same: using or combining finer features from lower layers and coarse semantic features for higher layers. For example,~\cite{sermanet2013pedestrian,bell2015inside,farabet2013learning,hariharan2015hypercolumns} combine features from multiple layers for the final classifier; while~\cite{sermanet2013pedestrian,bell2015inside} use subsampled features from finer scales,~\cite{farabet2013learning,hariharan2015hypercolumns} upsample the features to the finest scale and use their combination. Instead of combining features,~\cite{long2015fully,xie2015holistically,liu2015parsenet} do independent predictions at multiple layers and average the results.
In our proposed framework, such upsampling, subsampling and fusion operations can be easily controlled by the lateral and top-down connections.

The proposed TDM network is conceptually similar to the strategies explored in other contemporary works~\cite{pinheiro2016learning,ranjan2016optical,badrinarayanan2015segnet,lin2016feature,ronneberger2015unet}. All methods, including ours, propose architectures that go beyond the standard skip-connection paradigm and/or follow a coarse-to-fine strategy when using features from multiple levels of the bottom-up feature hierarchy. However, different methods focus on different tasks which guide their architectural design and training methodology.

Conv-deconv~\cite{noh2015learning} and encoder-decoder style networks~\cite{badrinarayanan2015segnet,ronneberger2015unet} have been used for image segmentation to utilize finer features via lateral connections. These connections generally use the `unpool' operation~\cite{zeiler2010deconvolutional}, which merely inverts the spatial pooling operation. Moreover, there is no modulation of bottom-up network. In comparison, our formulation is more generic and is responsible for the flow of high-level context features~\cite{piech2013network}.

Pinheiro \etal~\cite{pinheiro2016learning} focus on refining class-agnostic object proposals by first selecting proposals \emph{purely} based on bottom-up feedforward features~\cite{pinheiro2015learning}, and then \emph{post-hoc} learning how to refine each proposal independently using top-down and lateral connections (due to the computational complexity, only a few proposals can be selected to be refined). We argue that this use of top-down and lateral connections for refinement is sub-optimal for detection because it relies on the proposals selected based on feedforward features, which are insufficient to represent small and difficult objects. This training methodology also limits the ability to update the feedforward network through lateral connections. In contrast to this, we propose to learn better features for recognition tasks in an end-to-end trained system, and these features are used for both proposal generation and object detection. Similar to the idea of coarse-to-fine refinement, Ranjan and Black~\cite{ranjan2016optical} propose a coarse-to-fine spatial pyramid network, 
which computes a low resolution residual optical flow and iteratively improves predictions with finer pyramid levels. 
This is akin to \emph{only} using a specialized top-down network, which is suited for low-level tasks (like optical flow) but not for recognition tasks. 
Moreover, such purely coarse-to-fine networks cannot utilize models pre-trained on large-scale datasets~\cite{imagenet}, which is important for recognition tasks~\cite{rcnn}.
Therefore, our approach learns representation using bottom-up (fine-to-coarse), top-down (coarse-to-fine) and lateral networks simultaneously, and can use different pre-trained modules.

The proposed top-down network is closest to the recent work of Lin \etal~\cite{lin2016feature}, developed concurrently to ours, on feature pyramid network for object detection. Lin \etal~\cite{lin2016feature} use bottom-up, top-down and lateral connections to learn a feature pyramid, and require multiple proposal generators and region classifiers on each level of the pyramid. In comparison, the proposed top-down modulation focuses on using these connections to learn a single final feature map that is used by a single proposal generator and region classifier.

There is strong evidence of such top-down context, feedback and lateral processing in the human visual pathway~\cite{lamme1998feedforward,kravitz2013ventral,gilbert2007brain,lamme2000distinct,zanto2010top,zanto2011causal,gazzaley2012top,piech2013network,hopfinger2000neural,chun1999top}; wherein, the top-down signals are responsible for modulating low-level features~\cite{zanto2010top,zanto2011causal,gazzaley2012top,piech2013network,gilbert2007brain} as well as act as attentional mechanism for selection of features~\cite{hopfinger2000neural,chun1999top}. In this paper, we propose a computation model that captures some of these intuitions and incorporates them in a standard ConvNets, giving substantial performance improvements.

Our top-down framework is also related to the process of contextual feedback~\cite{bachmann2015hidden}. To incorporate top-down feedback loop in ConvNets, contemporary works~\cite{ief,gatta2014unrolling,li2015iterative,shrivastava2016contextual}, have used `unrolled' networks (trained stage-wise).
The proposed top-down network with lateral connections explores a complementary paradigm and can be readily combined with them. Contextual features have also been used for ConvNets based object detectors; \eg, using other objects~\cite{gupta2015exploring} or regions~\cite{gkioxari2015contextual} as context. We believe the proposed top-down path can naturally transmit these contextual features.

\vspace{-0.01in}
\section{Top-Down Modulation (TDM)}\label{sec:approach}
\vspace{-0.01in}

Our goal is to incorporate \emph{top-down modulation} into current object detection frameworks. The key idea is to select/attend to fine details from lower level feature maps based on top-down contextual features and select top-down contextual features based on the fine low-level details. We formalize this by proposing a simple top-down modulation (TDM) network  as shown in Figure~\ref{fig:overview_long}. 

The TDM network starts from the last layer of bottom-up feedforward network. For example, in the case of VGG16, the input to the first layer of the TDM network is the \conv\verb|5_3| output. Every layer of TDM network also gets the bottom-up features as inputs via lateral connections. Thus, the TDM network learns to: (a) transmit high-level contextual features that guide the learning and selection of relevant low-level features, and (b) use the bottom-up features to select the contextual information to transmit. The output of the proposed network captures both pertinent finer details and high-level information.

\vspace{-0.02in}
\subsection{Proposed Architecture}\label{sec:arch_details}
\begin{figure}[t]
    \centering
    \includegraphics[width=0.65\linewidth]{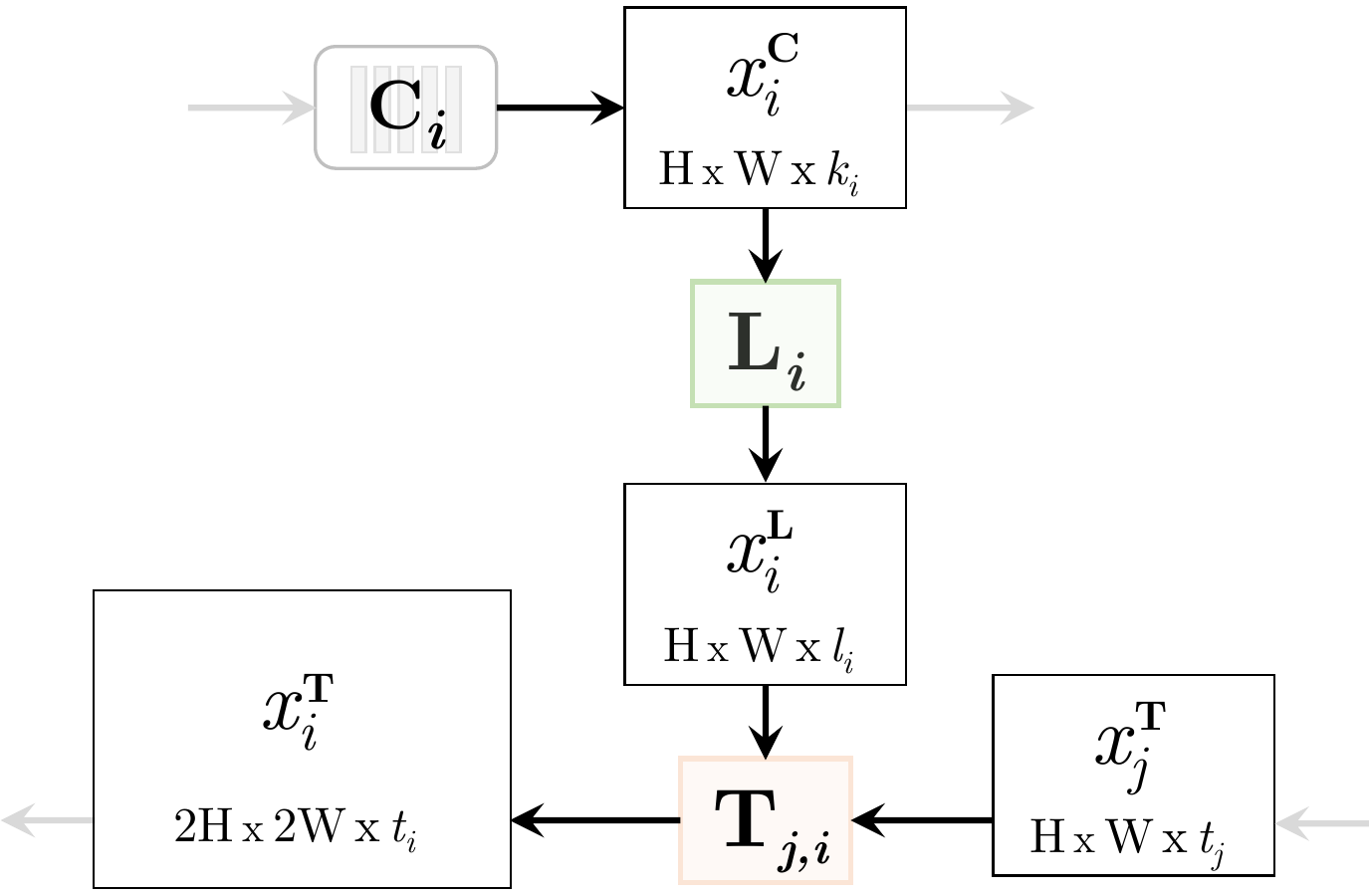}
    \caption{The basic building blocks of Top-Down Modulation Network (detailed Section~\ref{sec:arch_details}).
    }
    \label{fig:overview_short}
    \vspace{-0.1in}
\end{figure}

\vspace{-0.01in}
An overview of the proposed framework is illustrated in Figure~\ref{fig:overview_long}.
The standard bottom-up network is represented by blocks of layers, where each block $\mathbf{C}_i$ has multiple operations. The TDM network hinges on two key components: a lateral module $\mathbf{L}$, and a top-down module $\mathbf{T}$ (see Figure~\ref{fig:overview_short}). Each lateral module $\mathbf{L}_i$ takes in a bottom-up feature $x^\mathbf{C}_i$ (output of $\mathbf{C}_i$) and produces the corresponding lateral feature $x^\mathbf{L}_i$. These lateral features $x^\mathbf{L}_i$ and top-down features $x^\mathbf{T}_{j}$ are combined, and optionally upsampled, by the $\mathbf{T}_{j,i}$ module to produce the top-down features $x^\mathbf{T}_i$. These modules, $\mathbf{T}_{j,i}$ and $\mathbf{L}_i$, control the capacity of the modulation network by changing their output feature dimensions. 

The feature from the last top-down module $\mathbf{T}^\text{out}_i$ is used for the task of object detection. For example, in Figure~\ref{fig:overview_long}, instead of $x^\mathbf{C}_5$, we use $\mathbf{T}^\text{out}_2$ as input to ROI proposal and ROI classifier networks of the Faster R-CNN~\cite{ren2015faster} detection system (discussed in Section~\ref{sec:frrcn}). During training, gradient updates from the object detector backpropagate via top-down and lateral modules to the $\mathbf{C}_i$ blocks. The lateral modules $\mathbf{L}_\bullet$ learn how to transform low-level features and the top-down modules $\mathbf{T}_{\bullet}$ learn what semantic or context information to preserve in the top-down feature transmission as well as the selection of relevant low-level lateral features. Ultimately, the bottom-up features are modulated to adapt for this new representation. 

\vspace{-0.15in}
\paragraph{Architecture details.} The top-down and lateral modules described above are essentially small ConvNets, which can vary from a single or a hierarchy of convolutional layers to more involved blocks with Residual~\cite{resnet} or Inception~\cite{szegedy2016inception} units. In this paper, we limit our study by using modules with a single convolutional layer and non-linearity to analyze the impact of top-down modulation.

\begin{figure}[t]
    \centering
    \includegraphics[width=\linewidth]{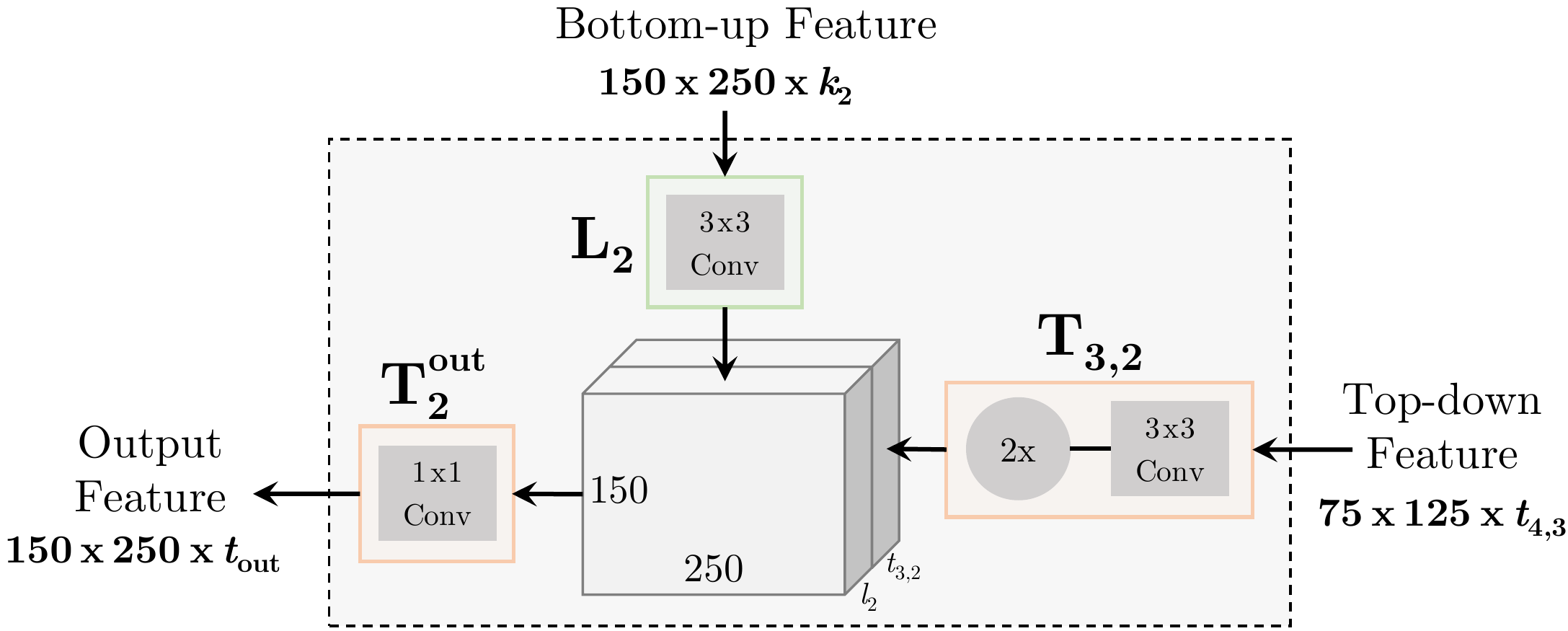}
    \caption{An example with details of top-down modules and lateral connections. Please see Section~\ref{sec:arch_details} for details of the architecture.}
    \vspace{-0.1in}
    \label{fig:overview_details}
\end{figure}

A detailed example of lateral and top-down modules is illustrated in Figure~\ref{fig:overview_details}. The lateral module $\mathbf{L}_i$ is a $3\times3$ convolutional layer with ReLU non-linearity, which transforms an $\left(\texttt{H}_i\times\texttt{W}_i\times k_i\right)$ input $x^\mathbf{C}_i$, to $\left(\texttt{H}_i\times\texttt{W}_i\times l_i\right)$ lateral feature $x^\mathbf{L}_i$. The top-down module $\mathbf{T}_{j,i}$ is also a $3\times3$ convolutional layer with ReLU, that combines this lateral feature with $\left(\texttt{H}_{i}\times\texttt{W}_{i}\times t_j\right)$ top-down feature $x^\mathbf{T}_{j}$, to produce an intermediate output $\left(\texttt{H}_{i}\times\texttt{W}_{i}\times t_{i}\right)$. If the resolution of next lateral feature $x^\mathbf{L}_{i-1}$ is higher than the previous lateral feature (\eg, $\texttt{H}_{i-1}= 2\times\texttt{H}_{i})$, then $\mathbf{T}_{j,i}$ also upsamples the intermediate output to produce $\left(\texttt{H}_{i-1}\times\texttt{W}_{i-1}\times t_{i}\right)$ top-down feature $x^\mathbf{T}_{i}$. In Figure~\ref{fig:overview_long}, we denote $a_i = t_j+l_i$ for simplicity. The final $\mathbf{T}^\text{out}_{i}$ module can additionally have a $1\times1$ convolutional layer with ReLU to output a $\left(\texttt{H}_i\times\texttt{W}_i\times k_{\text{out}}\right)$ feature, which is used by the detection system. 

Varying $l_i$, $t_i$ and $k_\text{out}$ controls the capacity of the top-down modulation system and dimension of the output features. These hyperparameters are governed by the base network design, detection system and hardware constraints (discussed in Section~\ref{sec:tdm_details}). Notice that the upsampling step is optional and depends on the content and arrangement of $\mathbf{C}$ blocks (\eg, no upsampling by $\mathbf{T}_4$ in Figure~\ref{fig:overview_long}). Also, the first top-down module ($\mathbf{T}_5$ in the illustration) only operates on $x^\mathbf{C}_5$ (the final output of the bottom-up network). 

\vspace{-0.1in}
\paragraph{Training methodology.} Integrating top-down modulation framework into a bottom-up ConvNet is only meaningful when the latter can represent high-level concepts in higher layers. Thus, we typically start with a pre-trained bottom-up network (see Section~\ref{sec:pretraining} for discussion). Starting with this pre-trained network, we find that progressively building the top-down network performs better in general. Therefore, we add one new pair of lateral and top-down modules at a time. For example, for the illustration in Figure~\ref{fig:overview_long}, we will begin by adding $\left(\mathbf{L}_4, \mathbf{T}_{5,4}\right)$ and use $\mathbf{T}^\text{out}_4$ to get features for object detection. After training $\left(\mathbf{L}_4, \mathbf{T}_{5,4}\right)$ modules, we will add the next pair $\left(\mathbf{L}_3, \mathbf{T}_{4,3}\right)$ and use a new $\mathbf{T}^\text{out}_3$ module to get features for detection; and we will repeat this process. With each new pair, the entire network, top-down and bottom-up along with lateral connections, is trained end-to-end. Implementation details of this training methodology depends on the base network architecture, and will be discussed in Section~\ref{sec:tdm_details}.

To better understand the impact of the proposed TDM network, we conduct extensive experimental evaluation; and provide ablation analysis of various design decisions. We describe our approach in detail (including preliminaries and implementation details) in Section~\ref{sec:details} and present our results in Section~\ref{sec:results}. We also report ablation analysis in Section~\ref{sec:analysis}. We would like to highlight that the proposed framework leads to substantial performance gains across different base network architectures, indicating its wide applicability.

\section{Approach Details}\label{sec:details}

\begin{table*}
    \centering
    \footnotesize
    \caption{Base networks architecture details for VGG16, ResNet101 and InceptionResNetv2. Legend: $\mathbf{C}_i$: bottom-up block id, N: number of convolutional filters, NR: number of residual units, NI: number of inception-resnet units, $\texttt{dim}\left(x^\mathbf{C}_i\right)$: Resolution and dimensions of the output feature. Refer to~\cite{VGG,resnet,szegedy2016inception,huang2016speed} for details\vspace{0.05in}}
    \label{tab:base_arch}
        \begin{tabular}{@{}lcrr@{}}
        \toprule
        \multicolumn{4}{c}{VGG16}\\
        \arrayrulecolor{gray}
        \midrule
        \arrayrulecolor{black}
        name & $\mathbf{C}_i$ & N & $\texttt{dim}\left(x^\mathbf{C}_i\right)$\\
        \midrule
        \conv1\_x & $\mathbf{C}_1$ & 2 & $\left(300,500,64\right)$\\
        \conv2\_x & $\mathbf{C}_2$ & 2 & $\left(150,250,128\right)$\\
        \conv3\_x & $\mathbf{C}_3$ & 3 & $\left(75,125,256\right)$\\
        \conv4\_x & $\mathbf{C}_4$ & 3 & $\left(37,63,512\right)$\\
        \conv5\_x & $\mathbf{C}_5$ & 3 & $\left(37,63,512\right)$\\
     \bottomrule
        \end{tabular}
    \quad
    \begin{tabular}{@{}lcrrr@{}}
    \toprule
    \multicolumn{5}{c}{ResNet101}\\
    \arrayrulecolor{gray}
    \midrule
    \arrayrulecolor{black}
    name & $\mathbf{C}_i$ & NB & N & $\texttt{dim}\left(x^\mathbf{C}_i\right)$\\
    \midrule
    \conv1 & $\mathbf{C}_1$ & 1 & 1 & $\left(300,500,64\right)$\\
    \conv2\_x & $\mathbf{C}_2$ & 3 & 9 & $\left(150,250,256\right)$\\
    \conv3\_x & $\mathbf{C}_3$ & 4 & 12 & $\left(75,125,512\right)$\\
    \conv4\_x & $\mathbf{C}_4$ & 23 & 69 & $\left(75,125,1024\right)$\\
         \bottomrule
         \\
    \end{tabular}
    \quad
    \begin{tabular}{@{}llrrr@{}}
    \toprule
    \multicolumn{5}{c}{InceptionResNetv2}\\
    \arrayrulecolor{gray}
    \midrule
    \arrayrulecolor{black}
    name & $\mathbf{C}_i$ & NI & N & $\texttt{dim}\left(x^\mathbf{C}_i\right)$\\
    \midrule
    \conv\_x & $\mathbf{C}_\text{x}$ & -  & 5   & $\left(71,246,192\right)$\\
    Mixed\_5b & $\mathbf{M}_5$ & 1  & 7   & $\left(35,122,320\right)$\\
    Block\_10x & $\mathbf{B}_{10}$ & 10  & 70   & $\left(35,122,320\right)$\\
    Mixed\_6a & $\mathbf{M}_{6a}$ & 1  & 3   & $\left(33,120,1088\right)$\\
    Block\_20x & $\mathbf{B}_{20}$ & 20 & 100 & $\left(33,120,1088\right)$\\
    \bottomrule
    \end{tabular}
    \vspace{-0.07in}
\end{table*}

In this section, we describe the preliminaries and provide implementation details of our top-down modulation (TDM) network under various settings. We first give a a brief overview of the object detection system and the ConvNet architectures used throughout this paper.

\subsection{Preliminaries: Faster R-CNN}\label{sec:frrcn}
We use the Faster R-CNN~\cite{ren2015faster} framework as our base object detection system.
Faster R-CNN consists of two core modules: 1) ROI Proposal Network (RPN), which takes an image as input and proposes rectangular regions of interests (ROIs); and 2) ROI Classifier Network (RCN), which is a Fast R-CNN~\cite{girshick2015fast} detector that classifies these proposed regions and learns to refine ROI coordinates. Given an image, Faster R-CNN first uses a ConvNet to extract features that are shared by both RPN and RCN. RPN uses these features to propose candidate ROIs, which are then classified by RCN. The RCN network projects each ROI onto the shared feature map and performs the `ROI Pooling'~\cite{girshick2015fast,he2014spatial} operation to extract a fixed length representation.
Finally, this feature is used for classification and box regression. See~\cite{girshick2015fast,he2014spatial,ren2015faster,huang2016speed} for details.

Due to lack of support for recent ConvNet architectures~\cite{resnet,szegedy2016inception} in the Faster R-CNN framework, we use our implementation in Tensorflow~\cite{abadi2016tensorflow}. We follow the design choices outlined in~\cite{huang2016speed}.
In Section~\ref{sec:results}, we will provide performance numbers using both the released code~\cite{ren2015faster} as well as our implementation (which tends to generally perform better). We use the end-to-end training paradigm for Faster R-CNN for all experiments~\cite{ren2015faster}. Unless specified otherwise, all methods start with models that were pre-trained on ImageNet classification~\cite{imagenet,rcnn}. 

\vspace{-0.01in}
\subsection{Preliminaries: Base Network Architectures}
\vspace{-0.01in}
In this paper, we use three standard ConvNet architectures: VGG16~\cite{VGG}, ResNet101~\cite{resnet} and InceptionResNetv2~\cite{szegedy2016inception}. We briefly explain how they are incorporated in the Faster R-CNN framework (see~\cite{ren2015faster,resnet,huang2016speed} for details), and give a quick overview of these architectures with reference to the bottom-up blocks $\mathbf{C}$ from Section~\ref{sec:arch_details}.

We use the term `Base Network' to refer to the part of ConvNet that is shared by both RPN and RCN; and `Classifier Network' to refer to the part that is used as RCN. For VGG16~\cite{VGG,ren2015faster}, ConvNet till \conv\verb|5_3| is used as the base network, and the following two \fc\ layers are used as the classifier network. Similarly, for ResNet101~\cite{resnet,ren2015faster}, base network is the ConvNet till \conv\verb|4_x|, and classifier network is the \conv\verb|5_x| block (with $3$ residual units or $9$ convolutional layers). For InceptionResNet101v2~\cite{szegedy2016inception,huang2016speed}, ConvNet till the `Block\_20x' is used as the base network, and the remaining layers (`Mixed\_7a' and `Block\_9x', with a total of 11 inception-resnet units or 48 convolutional layers) are used as the classifier network. Following~\cite{huang2016speed}, we change the pooling stride of the penultimate convolutional block in ResNet101 and InceptionResNetv2 to $1$ to maintain spatial resolution, and use atrous~\cite{deeplab,liu2015parsenet} convolution to recover the original field-of-view. Properties of bottom-up blocks $\mathbf{C}_i$, including number of layers, the output feature resolution and feature dimension \etc are given in Table~\ref{tab:base_arch}.

\begin{table}[t]
\footnotesize
\centering
\caption{\textbf{Top-Down Modulation} network design for VGG16, ResNet101 and InceptionResNetv2. Notice that $t_\text{out}$ VGG16 is much smaller than 512, thus requiring fewer parameters in RPN and RCN modules. Also note that it is important to keep $t_\text{out}$ fixed for ResNet101 and InceptionResNetv2 in order to utilize pre-trained RPN and RCN modules\vspace{0.05in}}
\label{tab:tdm_capacity}
\begin{tabular}{@{}llrrr@{}}
\toprule
\multicolumn{5}{c}{VGG16} \\
\arrayrulecolor{gray}
\midrule
\arrayrulecolor{black}
$\mathbf{T}_{i,j}$ & $\mathbf{L}_{j}$ & $t_{i,j}$ & $l_j$ & $t_\text{out}$ \\
\midrule
$\mathbf{T}_{5,4}$ & $\mathbf{L}_{4}$ & 128 & 128 & 256 \\
$\mathbf{T}_{4,3}$ & $\mathbf{L}_{3}$ & 64 & 64 & 128 \\
$\mathbf{T}_{3,2}$ & $\mathbf{L}_{2}$ & 64 & 64 & 128 \\
$\mathbf{T}_{2,1}$ & $\mathbf{L}_{1}$ & 64 & 64 & 128 \\
\bottomrule
\\
\end{tabular}
\quad 
\begin{tabular}{@{}llrrr@{}}
\toprule
\multicolumn{5}{c}{ResNet101} \\
\arrayrulecolor{gray}
\midrule
\arrayrulecolor{black}
$\mathbf{T}_{i,j}$ & $\mathbf{L}_{j}$ & $t_{i,j}$ & $l_j$ & $t_\text{out}$ \\
\midrule
$\mathbf{T}_{4,3}$ & $\mathbf{L}_{3}$ & 128 & 128 & 1024 \\
$\mathbf{T}_{3,2}$ & $\mathbf{L}_{2}$ & 128 & 128 & 1024 \\
$\mathbf{T}_{2,1}$ & $\mathbf{L}_{1}$ & 32 & 32 & 1024 \\
\bottomrule
\\
\\
\end{tabular}
\begin{tabular}{@{}llrrr@{}}
\toprule
\multicolumn{5}{c}{InceptionResNetv2} \\
\arrayrulecolor{gray}
\midrule
\arrayrulecolor{black}
$\mathbf{T}_{i,j}$ & $\mathbf{L}_{j}$ & $t_{i,j}$ & $l_j$ & $t_\text{out}$ \\
\midrule
$\mathbf{T}_{\text{B}_{20},\text{6a}}$ & $\mathbf{L}_{6\text{a}}$ & 576 & 512 & 1088 \\
$\mathbf{T}_{6\text{a},5\text{b}}$ & $\mathbf{L}_{5\text{b}}$ & 512 & 256 & 1088 \\
\bottomrule
\end{tabular}
\vspace{-0.1in}
\end{table}

\subsection{Top-Down Modulation}\label{sec:tdm_details}
To add the proposed TDM network to the ConvNet architectures described above, we need to decide the extent of top-down modulation, the frequency of lateral connections and the capacity of $\mathbf{T}$, $\mathbf{L}$ and $\mathbf{T}^\text{out}$ modules. 
We try to follow these principles when making design decisions: (a) coarse semantic modules need larger capacity; (b) lateral and top-down connections should reduce feature dimensionality in order to force selection; and (c) the capacity of $\mathbf{T}^\text{out}$ should be informed by the Proposal (RPN) and Classifier (RCN) Network design. Finally, the hardware constraint that a TDM augmented ConvNet should fit on a standard GPU.

\begin{table*}[th]
\centering
\caption{Object detection results on the COCO benchmark. Different methods use different networks for region proposal generation (ROINet) and for region classification (ClsNet). Results for the top block (except Faster R-CNN\protect\raisebox{0.2ex}{$\star$}) were directly obtained from their respective publications~\protect\cite{ren2015faster,resnet,pinheiro2016learning,shrivastava2016contextual}. Faster R-CNN\protect\raisebox{0.2ex}{$\star$} was reproduced by us. Middle block shows our implementation of Faster R-CNN framework, which we use as our primary baseline. Bottom block presents the main results of TDM network, with current \protect\hl{state-of-the-art} single-model performance highlighted.}
\vspace{0.02in}
\renewcommand{\arraystretch}{1.2}
\renewcommand{\tabcolsep}{1.2mm}
\resizebox{\linewidth}{!}{
\begin{tabular}{
@{}o{2.7cm} o{1.3cm} o{1.3cm} o{2.5cm} o{2.5cm} 
!{\color{gray}\vrule} ccc
!{\color{gray}\vrule} ccc
!{\color{gray}\vrule} ccc
!{\color{gray}\vrule} ccc
@{}}
\Xhline{1pt}
\textbf{Method} & \textbf{train} & \textbf{test} & \textbf{ROINet} & \textbf{ClsNet} &
AP & AP$^{50}$ & AP$^{75}$ &
AP$^\text{S}$ & AP$^\text{M}$ & AP$^\text{L}$ &
AR$^1$ & AR$^{10}$ & AR$^{100}$ &
AR$^\text{S}$ & AR$^\text{M}$ & AR$^\text{L}$  \\
\Xhline{1pt}
Faster R-CNN~\cite{ren2015faster} & train & val & VGG16 & VGG16 & 21.2 & 41.5 & - & -& -& -& -& -& -& -& -& -\\
Faster R-CNN~\cite{resnet} & train & val & ResNet101 & ResNet101 & 27.2 & 48.4 & - & -& -& -& -& -& -& -& -& -\\
SharpMask~\cite{pinheiro2016learning} & train & testdev & ResNet50 & VGG16 & 25.2 & 43.4 & - & -& -& -& -& -& -& -& -& -\\
Faster R-CNN~\cite{ren2015faster}\raisebox{0.2ex}{$\star$} & trainval & testdev & VGG16 & VGG16 & 
24.5 & 46.0 & 23.7 & 8.2 & 26.4 & 36.9 & 24.0 & 34.8 & 35.5 & 13.4 & 39.2 & 54.3 \\
~\cite{shrivastava2016contextual} & trainval & testdev & VGG16++ & VGG16++ & 27.5 & 49.2 & 27.8 & 8.9 & 29.5 & 41.5 & 25.5 & 37.4 & 38.3 & 14.6 & 42.5 & 57.4 \\
\Xhline{1pt}
Faster R-CNN & trainval\raisebox{0.2ex}{$\ast$} & testdev & VGG16 & VGG16 & 23.3 & 44.7 & 21.5 & 9.4 & 27.1 & 32.0 & 22.7 & 36.8 & 39.4 & 18.3 & 44.0 & 56.2 \\
Faster R-CNN & trainval\raisebox{0.2ex}{$\ast$} & testdev & ResNet101 & ResNet101 & 31.5 & 52.8 & 33.3 & 13.6 & 35.4 & 44.5 & 28.0 & 43.6 & 45.8 & 22.7 & 51.2 & 64.1\\
Faster R-CNN & trainval\raisebox{0.2ex}{$\ast$} & testdev & IRNv2 & IRNv2 & 34.7 & 55.5 & 36.7 & 13.5 & 38.1 & 52.0 & 29.8 & 46.2 & 48.9 & 23.2 & 54.3 & 70.8 \\
\Xhline{1pt}
TDM [ours]  & trainval\raisebox{0.2ex}{$\ast$} & testdev & VGG16 + TDM & VGG16 + TDM & 28.6 &  48.1 &  30.4 &  14.2 &  31.8 &  36.9 &  26.2 &  42.2 &  44.2 &  23.7 &  48.3 &  59.3 \\
TDM [ours]  & trainval\raisebox{0.2ex}{$\ast$} & testdev & ResNet101 + TDM & ResNet101 + TDM & 35.2 & 55.3 & 38.1 & 16.6 & 38.4 & 47.9 & 30.4 & 47.8 & 50.3 & 27.8 & 54.9 & 67.6 \\
TDM [ours]  & trainval\raisebox{0.2ex}{$\ast$} & testdev & IRNv2 + TDM & IRNv2 + TDM & \hl{37.3} & \hl{57.8} & \hl{39.8} & \hl{17.1} & \hl{40.3} & \hl{52.1} & \hl{31.6} & \hl{49.3} & \hl{51.9} & \hl{28.1} & \hl{56.6} & \hl{71.1} \\
\Xhline{1pt}
\end{tabular}
}
\vspace{-0.11in}
\label{tab:coco_pref}
\end{table*}

To build a TDM network, we start with a standard bottom-up model trained on the detection task, and add $\left(\mathbf{T}_{i,j}, \mathbf{L}_j\right)$ progressively. The capacity for different $\mathbf{T}$, $\mathbf{L}$, and $\mathbf{T}^\text{out}$ modules is given in Table~\ref{tab:tdm_capacity}. For the VGG16 network, we add top-down and lateral modules all the way to the \conv\verb|1_x| feature. Notice that the input feature dimension to RPN and RCN networks changes from 512 (for \conv\verb|5_x|) to 256 (for $\mathbf{T}^\text{out}_{4}$), therefore we initialize the \fc\ layers in RPN and RCN randomly~\cite{xavier}. However, since $t_\text{out}$ is same for the last three $\mathbf{T}^\text{out}$ modules, we re-use the RPN and RCN layers for these modules.

For the ResNet101 and InceptionResNetv2, we add top-down and lateral modules for till \conv\verb|1| and `Mixed\_5b' respectively. Similar to VGG16, their base networks are initialized with a model pre-trained on the detection task. However, as opposed to VGG16, where the RCN has just 2 \fc\ layers, ResNet101 and InceptionResNetv2 models have an RCN with 9 and 48 convolutional layers respectively. This makes training RCN from random initialization difficult. To counter this, we ensure that all $\mathbf{T}^\text{out}$ output feature dimensions ($t_\text{out}$) are same, so that we can be readily use pre-trained RPN and RCN. This is implemented using an additional $1\times1$ convolutional layer wherever $\left(t_{i,j}+l_j\right)$ differs from $t_\text{out}$ (\eg, all $\mathbf{T}^\text{out}$ modules in ResNet101, and the final $\mathbf{T}^\text{out}$ module in InceptionResNetv2).

We would like to highlight an issue with training of RPN at high-resolution feature maps. RPN is a fully convolutional module of Faster R-CNN, that generates an intermediate 512 dimension representation which is of the same resolution as input; and losses are computed at all pixel locations. This is efficient for coarse features (\eg, last row in Table~\ref{tab:base_arch}), but the training becomes prohibitively slow for finer resolution features. To counter this, we apply RPN at a stride which ensures that computation remains exactly the same (\eg, using stride of 8 for $\mathbf{T}^\text{out}_{1}$ in VGG16). Because of `ROI Pooling' operations, RCN module still efficiently utilizes the finer resolution features.

\vspace{-0.01in}
\section{Results}\label{sec:results}
\vspace{-0.01in}

In this section, we evaluate our method on the task of object detection, and demonstrate consistent and significant improvement in performance when using features from the proposed TDM network.

\vspace{-0.1in}
\paragraph{Dataset and metrics.} All experiments and analysis in this paper are performed on the COCO dataset~\cite{lin2014coco}. All models were trained on 40k train and 32k val images (which we refer to as `\trainval' set). All ablation evaluations were performed on 8k val images (`\minival' set) held out from the val set. We also report quantitative results on the standard testdev2015 split. For quantitative evaluation, we use the COCO evaluation metric of mean average precision (AP\footnote{COCO~\cite{lin2014coco} AP averages over classes, recall, and IoU levels.}). 

\vspace{-0.1in}
\paragraph{Experimental Setup.} We conduct experiments with three standard ConvNet architectures: VGG16~\cite{VGG}, ResNet101~\cite{resnet} and InceptionResNetv2~\cite{szegedy2016inception}. All models (`Baseline' Faster R-CNN and ours) were trained with SGD for 1.5M mini-batch iterations, with batch size of 256 and 128 ROIs for RPN and RCN respectively. We start with an initial learning rate of 0.001 and decay it by 0.1 at 800k and 900k iterations. 

\vspace{-0.1in}
\paragraph{Baselines.} Our primary baseline is using vanilla VGG16, ResNet101 and InceptionResNetv2 features in the Faster R-CNN framework. However, due to lack of implementations supporting all three ConvNets, we opted to re-implement Faster R-CNN in Tensorflow~\cite{abadi2016tensorflow}. The baseline numbers reported in Table~\ref{tab:coco_pref}(middle) are using our implementation and training schedule and are generally higher than the ones reported in~\cite{ren2015faster,shrivastava2016contextual}. Faster R-CNN\raisebox{0.2ex}{$\star$} was reproduced by us using the official implementation~\cite{ren2015faster}. All other results in  Table~\ref{tab:coco_pref}(top) were obtained from the original papers. 

We also compare against models which use a single region proposal generator and a single region classifier network. In particular, we compare with SharpMask~\cite{pinheiro2016learning}, because of its refinement modules with top-down and lateral connections, and~\cite{shrivastava2016contextual} because they also augment the standard VGG16 network with top-down information. Note that different methods use different networks and train/test splits (see Table~\ref{tab:coco_pref}), making it difficult to do a comprehensive comparison. Therefore, for discussion, we will directly compare against our Faster R-CNN baseline (Table~\ref{tab:coco_pref}(middle)), and highlight that the improvements obtained by our approach are much bigger than the boosts by other methods.
\subsection{COCO Results}
In Table~\ref{tab:coco_pref}(bottom), we report results of the proposed TDM network on the testdev2015 split of the COCO dataset. We see that the TDM network leads to a \textbf{5.3} AP point boost over the vanilla VGG16 network ($28.6$ AP \vs $23.3$ AP), indicating that TDM learns much better features for object detection. Note that even though our algorithm is trained on less data (\trainval), we outperform all methods with VGG16 architecture. For ResNet101, we improve the performance by  \textbf{3.7} points to 35.2 AP. InceptionResNetv2~\cite{szegedy2016inception} architecture was the cornerstone of the winning entry to COCO 2016 detection challenge~\cite{huang2016speed}. The best single model performance used by this entry achieves $34.7$ AP on the testdev split (\cite{huang2016speed}). Using InceptionResNetv2 as base, our TDM network achieves \textbf{37.3} AP, which is currently the \textbf{state-of-the-art}  single model performance on the testdev split without any bells and whistles (\eg, multi-scale, iterative box refinement, \etc). In fact, the TDM network outperforms the baselines (with same base networks) on almost all AP and AR metrics. Similarly, in Table~\ref{tab:coco_minival}, we observe that TDM achieves similar boosts across all network architectures on the \minival\ split as well.

\begin{figure}
    \centering
    \includegraphics[height=3.96in]{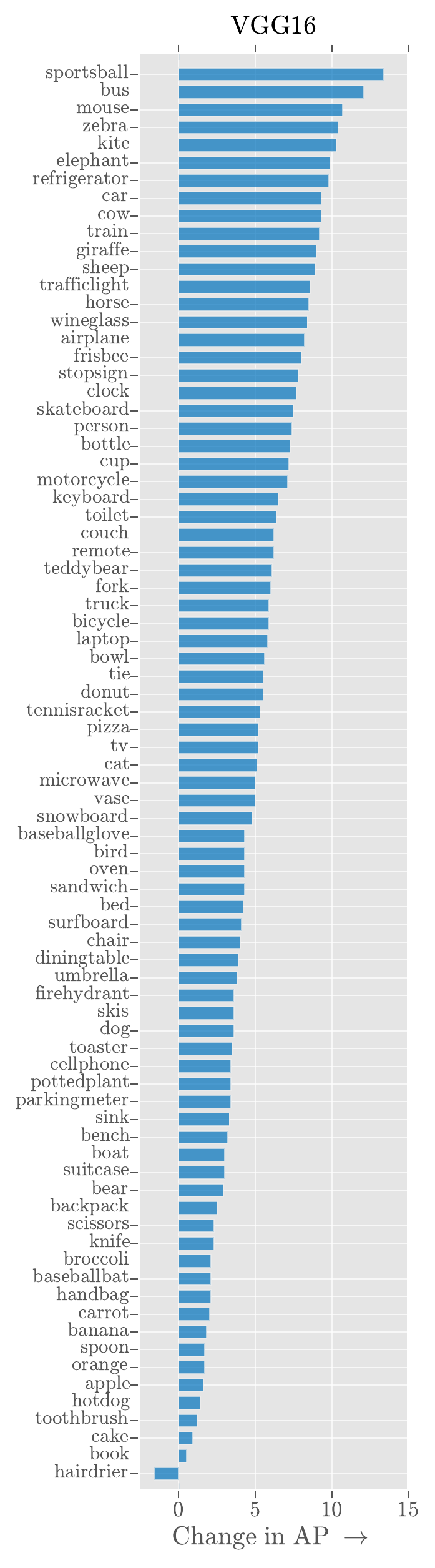} 
    \includegraphics[height=3.96in]{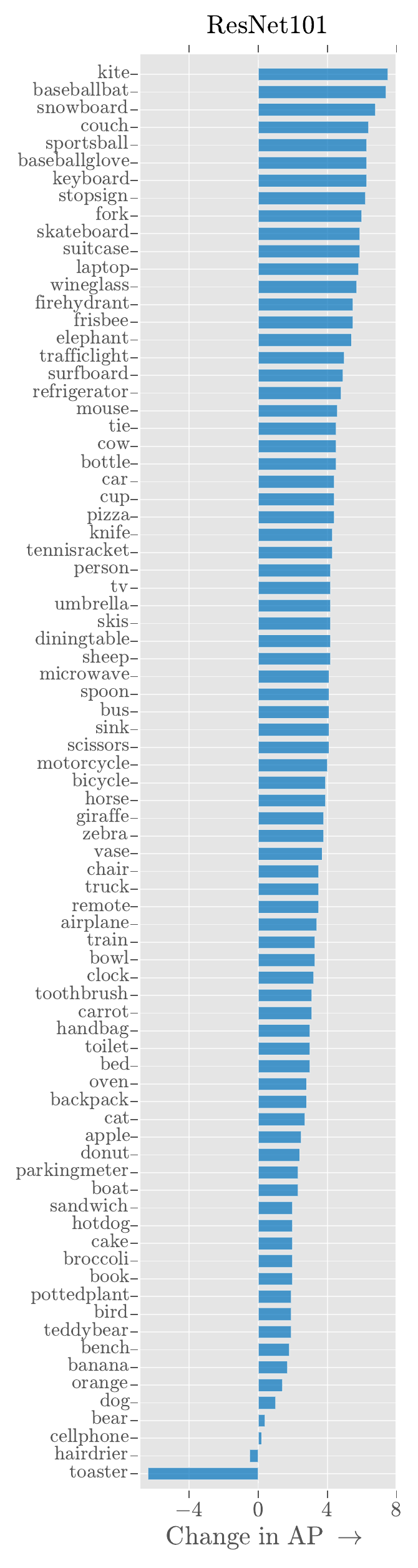} 
    \includegraphics[height=3.96in]{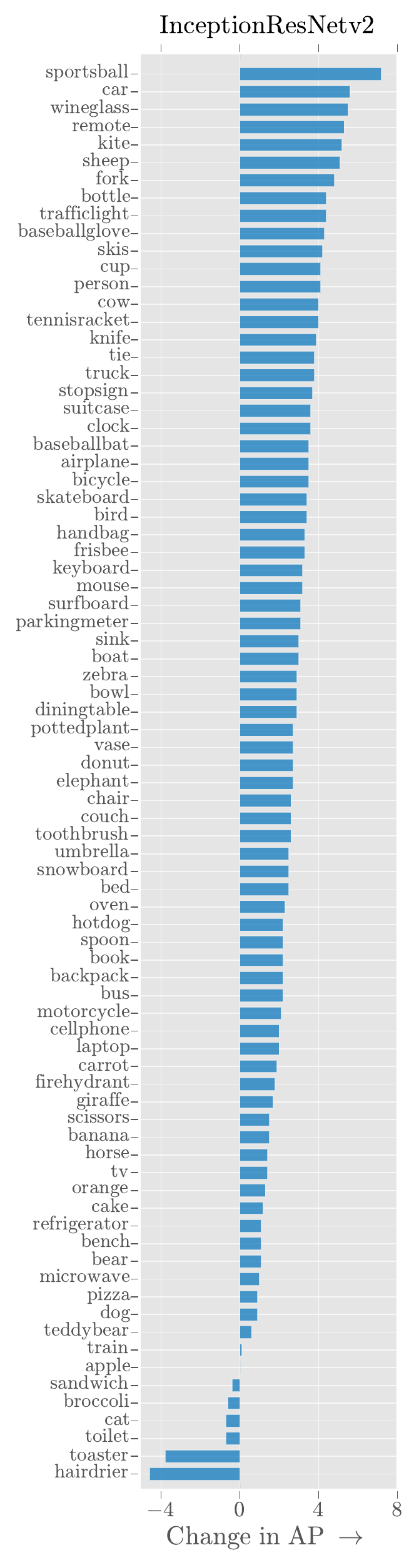} 
    \caption{Improvement in AP over Faster R-CNN baseline. Base Networks: (left) VGG16, (middle) ResNet101, and (right) InceptionResNetv2. Improved performance for almost all categories emphasize the effectiveness of Top-Down Modulation for object detection. (best viewed digitally)}
    \label{fig:distribution}
    \vspace{-0.1in}
\end{figure}

Figure~\ref{fig:distribution} shows change in AP from Faster R-CNN baseline to the TDM network (for the testdev split). When using VGG16, for all but one category, TDM features improve the performance on object detection. In fact, more than 50\% of the categories improve by 5 AP points or more, highlighting that the features are good for small and big objects alike. Similar trends hold for ResNet101 and InceptionResNetv2.

\vspace{-0.1in}
\paragraph{Improved localization.} In Table~\ref{tab:coco_pref}, we also notice that for VGG16, our method performs exceptionally well on the AP$^\text{75}$ metric, improving the baseline Faster R-CNN by \textbf{8.9} AP$^\text{75}$ points, which is much higher than the \textbf{3.5} point AP$^\text{50}$ boost. We believe that using contextual features to select and integrate low-level finer details is they key reason for this improvement. Similarly for ResNet101 and InceptionResNetv2, we see \textbf{4.8} AP$^\text{75}$ and \textbf{3.1} AP$^\text{75}$ boost respectively.

\vspace{-0.1in}
\paragraph{Improvement for small objects.} In Table~\ref{tab:coco_pref}, for VGG16, ResNet101 and InceptionResNetv2), we see \textbf{4.8}, \textbf{3} and \textbf{3.6} point boost respectively for small objects (AP$^\text{S}$) highlighting the effectiveness of features with TDM. Moreover, small objects are often on top of the list in Figure~\ref{fig:distribution} (\eg, sportsball +13 AP point, mouse +10 AP for VGG16). This is in line with other studies~\cite{gupta2015exploring}, which show that context is particularly helpful for some objects. Similar trends hold for the \minival\ split as well: \textbf{5.6}, \textbf{7.4} and \textbf{8.5} AP $^\text{S}$ boost for VGG16, ResNet101 and InceptionResNetv2 respectively.

\begin{table*}[t]
\centering
\caption{Ablation analysis on the COCO benchmark using the Faster R-CNN detection framework. All methods are trained on \protect\trainval\ and evaluated on \protect\minival\ set (Section \protect\ref{sec:results}). Methods are grouped based on their base network, \protect\hl{best} results are highlighted in each group.\vspace{0.02in}}
\renewcommand{\arraystretch}{1.2}
\renewcommand{\tabcolsep}{1.2mm}
\resizebox{\linewidth}{!}{
\footnotesize
\begin{tabular}{
@{}L{2.2cm} L{2.5cm} o{2.4cm}
!{\color{gray}\vrule} ccc
!{\color{gray}\vrule} ccc
!{\color{gray}\vrule} ccc
!{\color{gray}\vrule} ccc
@{}}
\Xhline{1pt}
\textbf{Method} & \textbf{Net} & \textbf{Features from:} & 
AP & AP$^{50}$ & AP$^{75}$ &
AP$^\text{S}$ & AP$^\text{M}$ & AP$^\text{L}$ &
AR$^1$ & AR$^{10}$ & AR$^{100}$ &
AR$^\text{S}$ & AR$^\text{M}$ & AR$^\text{L}$  \\
\Xhline{1pt}
Baseline & VGG16 & $\mathbf{C}_5$ & 25.5 & 46.7 & 24.6 & 6.1 & 23.3 & 37.0 & 23.9 & 38.2 & 40.7 & 14.1 & 39.5 & 55.3 \\
Skip-pool & VGG16 &$\mathbf{C}_2,\mathbf{C}_3,\mathbf{C}_4,\mathbf{C}_5$  & 25.3 & 46.3 & 25.9 & 9.1 & 24.0 & 36.0 & 24.6 & 40.0 & 42.4 & 18.6 & 41.8 & 54.1 
\\
TDM [ours] & VGG16 + TDM & $\mathbf{T}^\text{out}_{4}$ 
& 26.2 & 45.7 & 27.2 & 9.4 & 25.1 & 34.8 & 25.0 & 40.7 & 43.0 & 18.7 & 43.1 & 54.7 \\
TDM [ours] & VGG16 + TDM & $\mathbf{T}^\text{out}_{3}$  & 28.8 & 48.6 & 30.7 & 11.0 & 27.1 & 37.3 & 26.5 & 42.7 & 45.0 & 21.1 & 44.2 & 56.4 \\
TDM [ours] & VGG16 + TDM & $\mathbf{T}^\text{out}_{2}$ & \hl{29.9} & \hl{50.3} & 31.6 & 11.4 & \hl{28.1} & 38.6 & \hl{27.3} & \hl{43.7} & \hl{46.0} & 22.8 & 44.7 & \hl{57.1}  \\
TDM [ours] & VGG16 + TDM & $\mathbf{T}^\text{out}_{1}$  & 29.8 & 49.9 & \hl{31.7} & \hl{11.7} & 28.0 & \hl{39.3} & 27.1 & 43.5 & 45.9 & \hl{23.9} & \hl{45.4} & 56.8 \\
\Xhline{1pt}

Baseline & ResNet101 &  $\mathbf{C}_5$ & 32.1 & 53.2 & 33.8 & 9.4 & 29.7 & 45.7 & 28.3 & 44.3 & 46.7 & 19.3 & 46.3 & 60.9 \\
TDM [ours] & ResNet101 + TDM & $\mathbf{T}^\text{out}_{3}$ & 34.4 & 54.4 & 37.1 & 10.9 & 31.8 & 48.2 & 30.1 & 47.5 & 49.8 & 21.7 & 49.1 & 64.0 \\
TDM [ours] & ResNet101 + TDM & $\mathbf{T}^\text{out}_{2}$ & 35.3 & 55.1 & 38.3 & 11.2 & 33.0 & 48.2 & 30.7 & 48.0 & 50.5 & 22.5 & 50.1 & 63.6 \\
TDM [ours] & ResNet101 + TDM & $\mathbf{T}^\text{out}_{1}$ & \hl{35.7} & \hl{56.0} & \hl{38.5} & \hl{16.8} & \hl{39.2} & \hl{49.0} & \hl{30.9} & \hl{48.5} & \hl{50.9} & \hl{28.1} & \hl{55.6} & \hl{68.5} \\
\Xhline{1pt}
Baseline & IRNv2 & $\mathbf{B}_{20}$ & 35.7 & 56.5 & 38.0 & 8.9 & 32.0 & 52.5 & 30.8 & 47.8 & 50.3 & 19.6 & 49.9 & 66.9 \\
TDM [ours] & IRNv2 + TDM & $\mathbf{T}^\text{out}_{6a}$ & 37.3 & 57.9 & 39.5 & 11.4 & 33.3 & 53.3 & \hl{32.8} & 49.1 & 51.5 & 22.7 & 50.6 & 67.5 \\
TDM [ours] & IRNv2 + TDM & $\mathbf{T}^\text{out}_{5a}$ & \hl{38.1} & \hl{58.6} & \hl{40.7} & \hl{17.4} & \hl{41.1} & \hl{54.7} & 32.4 & \hl{50.1} & \hl{52.6} & \hl{28.9} & \hl{57.2} & \hl{72.3} \\
\Xhline{1pt}
\end{tabular}
}
\vspace{-0.1in}
\label{tab:coco_minival}
\end{table*}

\begin{figure*}
\centering
\includegraphics[height=1.58in,width=\maxwidth{1.58in},keepaspectratio]{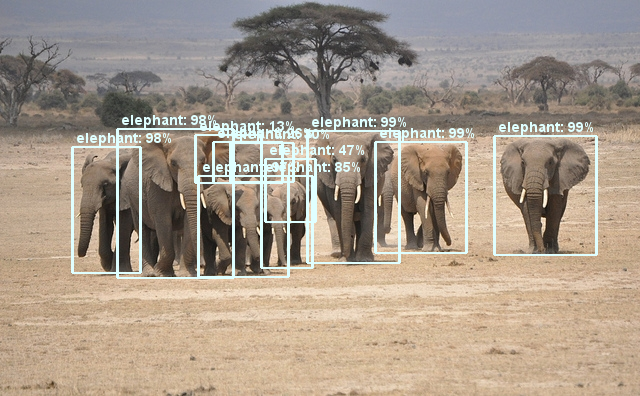}\hspace{0.1in}
\includegraphics[height=1.58in,width=\maxwidth{1.58in},keepaspectratio]{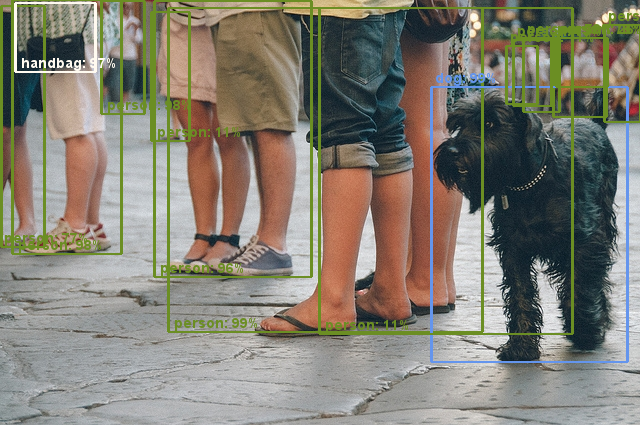}\hspace{0.1in}\vspace{0.1in}
\includegraphics[height=1.58in,width=\maxwidth{1.58in},keepaspectratio]{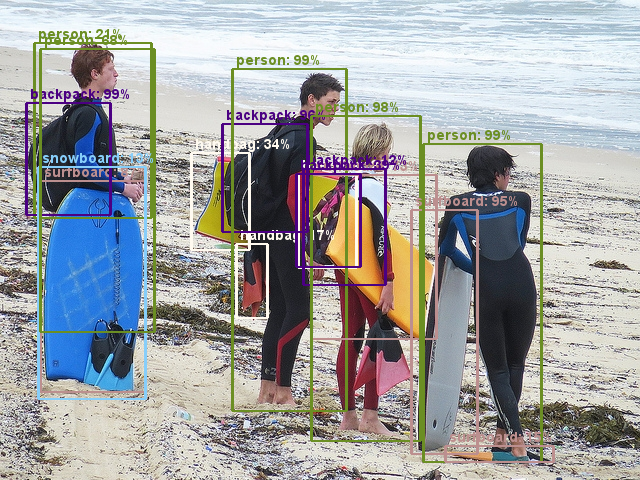}\hspace{0.1in}
\includegraphics[height=1.58in,width=\maxwidth{1.58in},keepaspectratio]{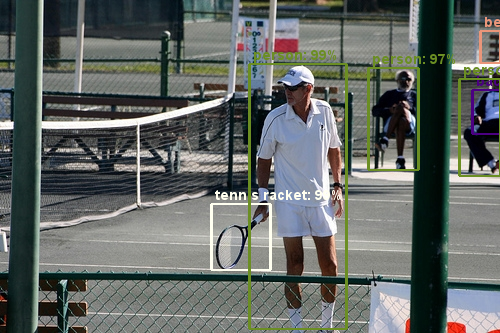}\\ 
\includegraphics[height=1.58in,width=\maxwidth{1.58in},keepaspectratio]{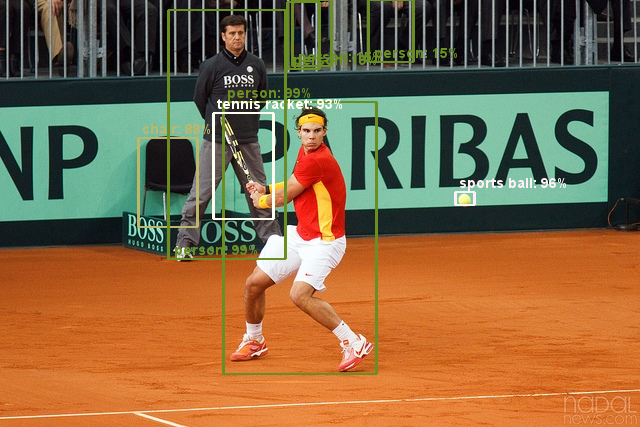}\hspace{0.1in}
\includegraphics[height=1.58in,width=\maxwidth{1.58in},keepaspectratio]{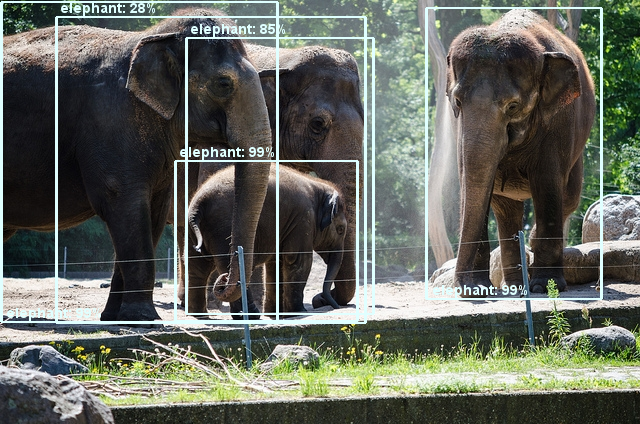}\hspace{0.1in}
\includegraphics[height=1.58in,width=\maxwidth{1.58in},keepaspectratio]{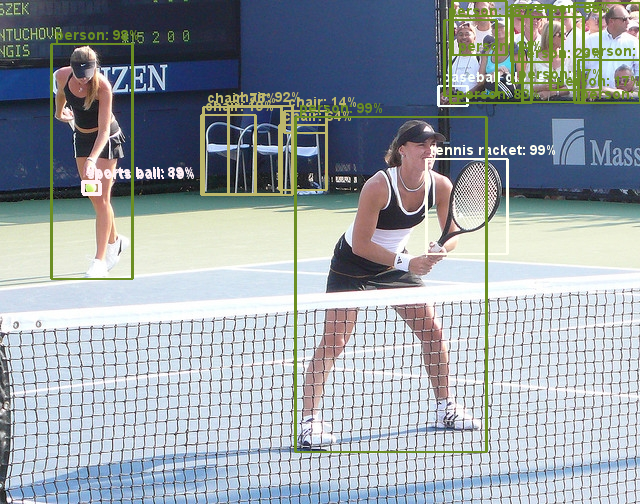}\hspace{0.1in}\vspace{0.1in}
\includegraphics[height=1.58in,width=\maxwidth{1.58in},keepaspectratio]{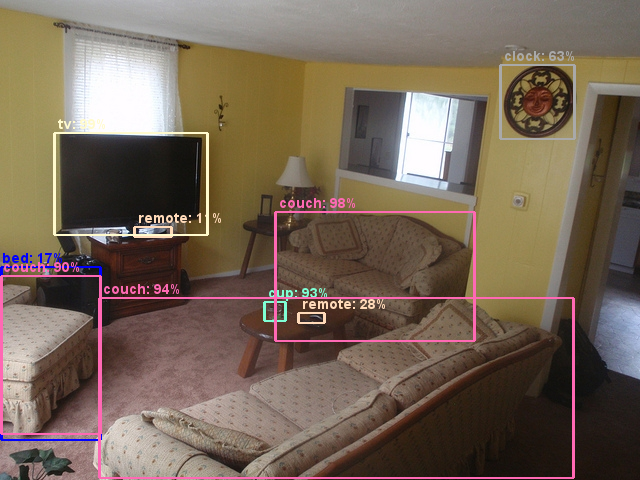}\hspace{0.1in}\\
\includegraphics[height=1.58in,width=\maxwidth{1.58in},keepaspectratio]{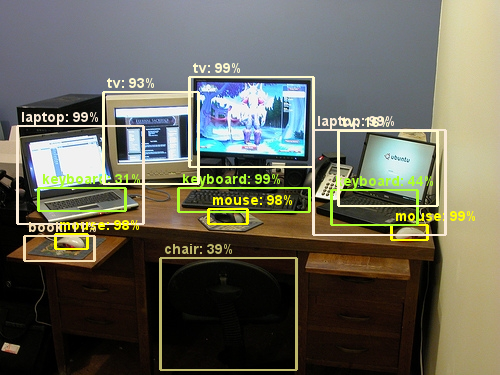}\hspace{0.1in}
\includegraphics[height=1.58in,width=\maxwidth{1.58in},keepaspectratio]{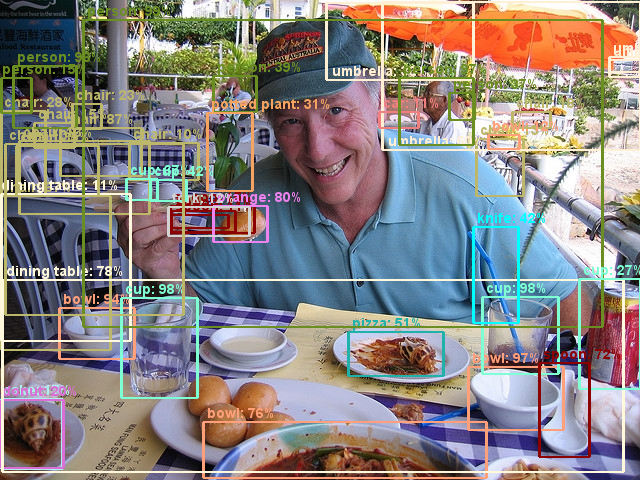}\hspace{0.1in}
\includegraphics[height=1.58in,width=\maxwidth{1.58in},keepaspectratio]{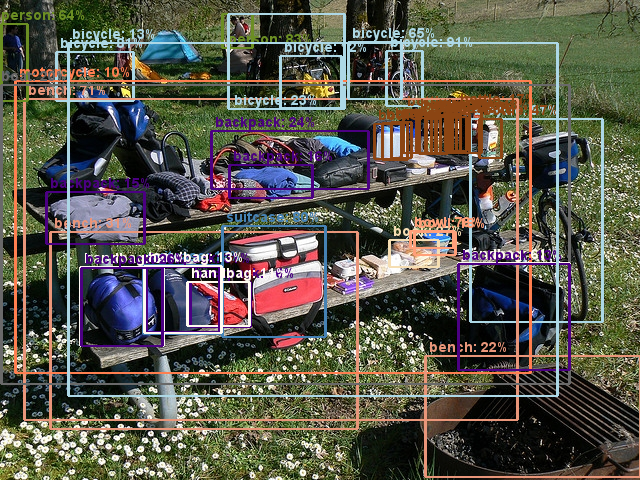}\hspace{0.1in}\vspace{0.1in}
\includegraphics[height=1.58in,width=\maxwidth{1.58in},keepaspectratio]{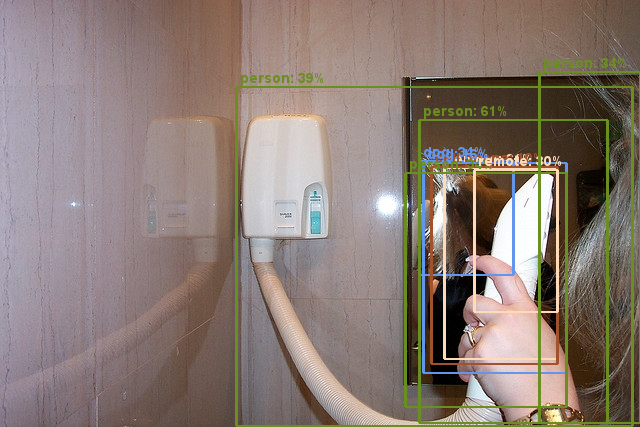}\\
\includegraphics[height=1.58in,width=\maxwidth{1.58in},keepaspectratio]{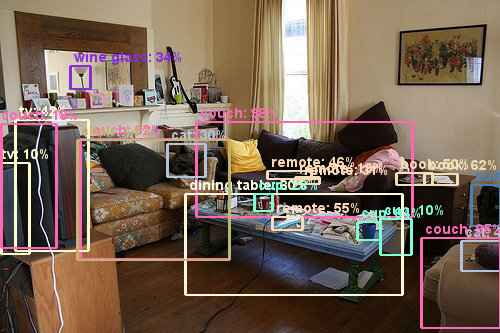}\hspace{0.1in}
\includegraphics[height=1.58in,width=\maxwidth{1.58in},keepaspectratio]{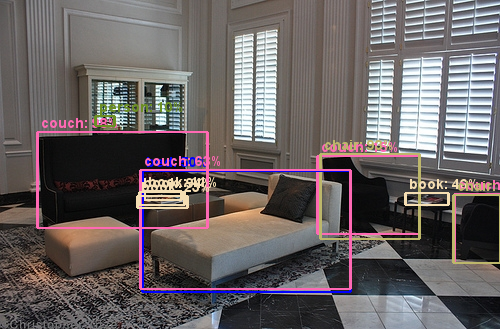}\hspace{0.1in}
\includegraphics[height=1.58in,width=\maxwidth{1.58in},keepaspectratio]{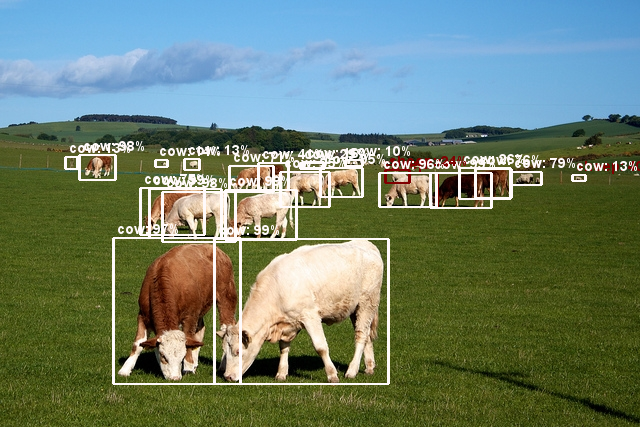}\hspace{0.1in}\vspace{0.1in}
\includegraphics[height=1.58in,width=\maxwidth{1.58in},keepaspectratio]{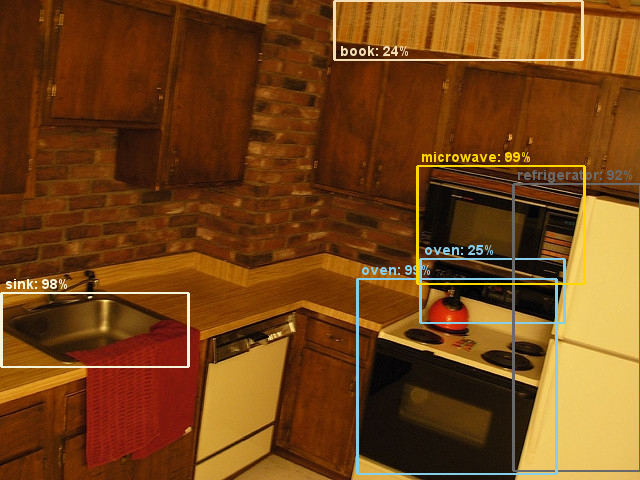}\\
\includegraphics[height=1.58in,width=\maxwidth{1.58in},keepaspectratio]{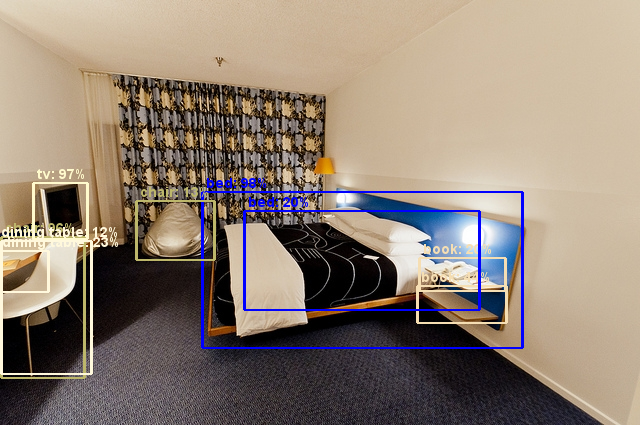}\hspace{0.1in}
\includegraphics[height=1.58in,width=\maxwidth{1.58in},keepaspectratio]{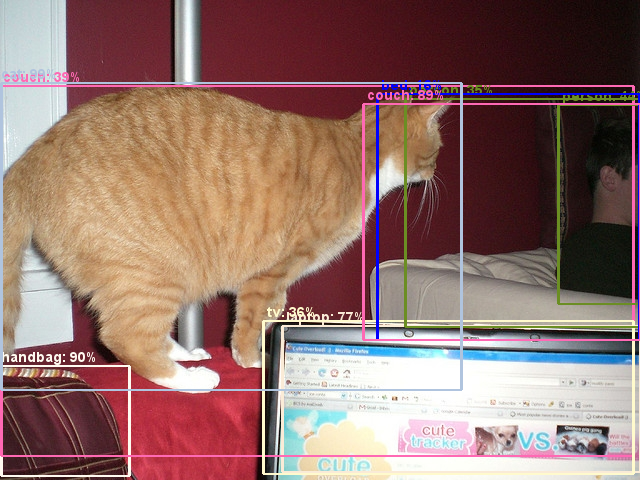}\hspace{0.1in}
\includegraphics[height=1.58in,width=\maxwidth{1.58in},keepaspectratio]{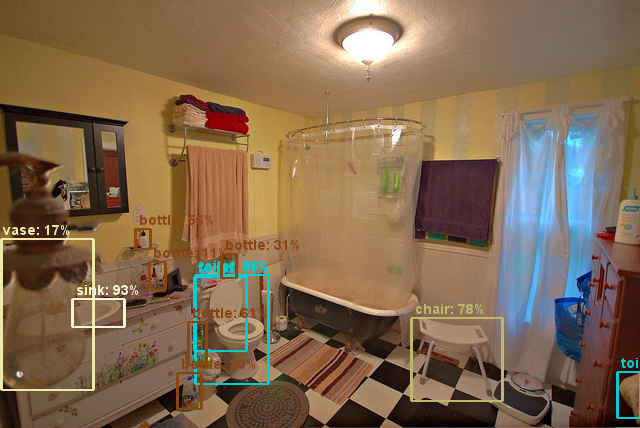}\hspace{0.1in}\vspace{0.1in}
\includegraphics[height=1.58in,width=\maxwidth{1.58in},keepaspectratio]{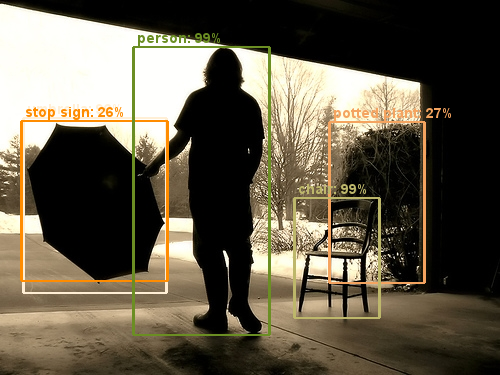}\\
\includegraphics[height=1.58in,width=\maxwidth{1.58in},keepaspectratio]{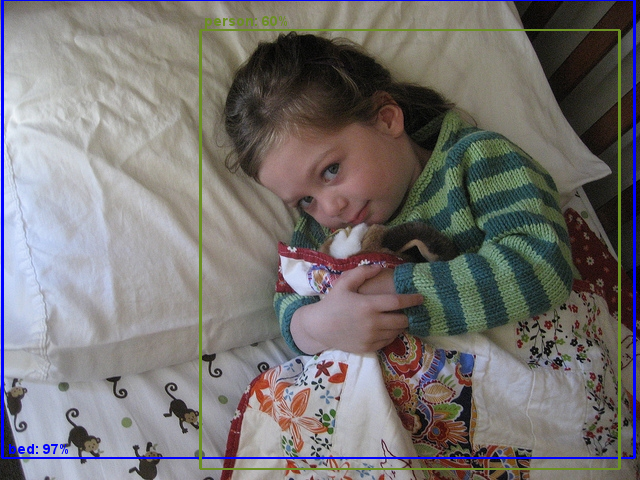}\hspace{0.1in}
\includegraphics[height=1.58in,width=\maxwidth{1.58in},keepaspectratio]{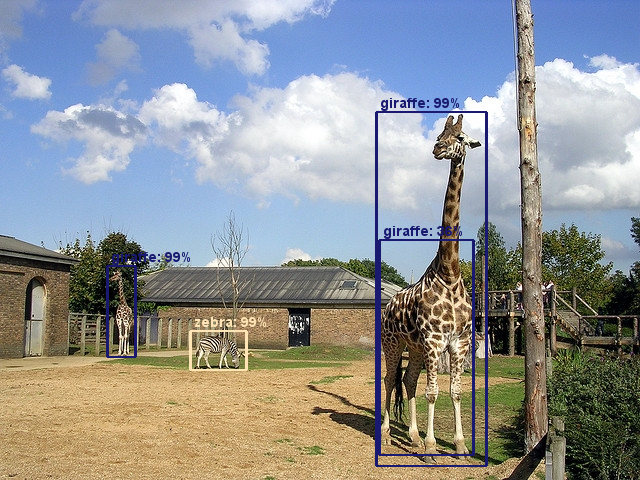}\hspace{0.1in}
\includegraphics[height=1.58in,width=\maxwidth{1.58in},keepaspectratio]{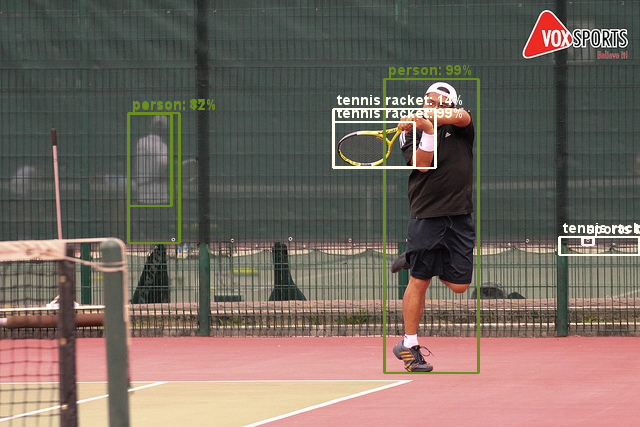}\hspace{0.1in}\vspace{0.1in}
\includegraphics[height=1.58in,width=\maxwidth{1.58in},keepaspectratio]{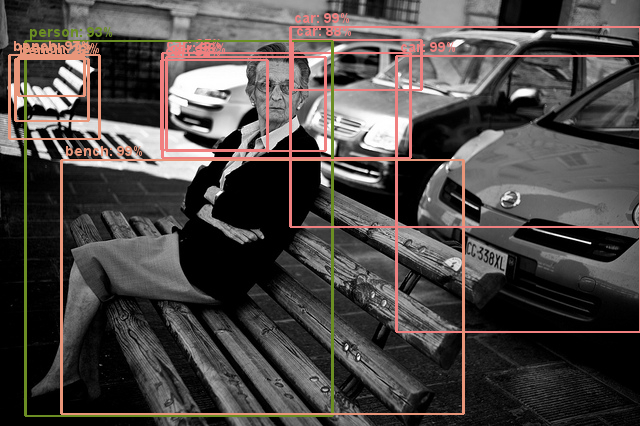}
\caption[{\small Qualitative results of the TDM network.}]{\textbf{Qualitative results} of the proposed TDM network on randomly selected images from the \minival\ set (best viewed digitally).}
\label{fig:tdm_qual}
\vspace{-0.1in}
\end{figure*}

\paragraph{Qualitative Results.}
In Figure~\ref{fig:tdm_qual}, we display qualitative results of the Top-down Modulation Network. Notice that the network can find small objects like remote (second row, last column; fourth row, first column) and sportsball (second row, third and fourth column). Also notice the detection results in the presence of heavy clutter.

\section{Design and Ablation Analysis}\label{sec:analysis}
In this section, we perform control and ablative experiments to study the importance of top-down and lateral modules in the proposed TDM network.

\subsection{How low should the Top-Down Modulation go?}\label{sec:howlow}
In Section~\ref{sec:tdm_details}, we discussed the principles that we follow to add top-down and lateral modules. We connected these modules till the lowest layer choosing design decisions that hardware constraints would permit. However, is that overkill? Is there an optimal layer, after which this modulation does not help or starts hurting? To answer these questions, we limit the layer till which the modulation process happens, and use those features for Faster R-CNN.

Recall that the TDN network is built progressively, \ie, we add one pair of lateral and top-down module at a time. So for this control study, we simply let each subsequent pair train for the entire learning schedule, treating the $\mathbf{T}^\text{out}$ as the final output. We report the results on \minival\ set in Table~\ref{tab:coco_minival}. As we can see, adding more top-down modulation helps in general. However, for VGG16, we see that the performance saturates at $\mathbf{T}^\text{out}_2$, and adding modules till $\mathbf{T}^\text{out}_1$ do not seem to help much. Deciding the endpoint criteria for top-down modulation is an interesting future direction.

\subsection{No lateral modules}
To analyze the importance of lateral modules, and to control for the extra parameters added by the TDM network (Table~\ref{tab:tdm_capacity}), we train additional baselines with variants of VGG16 $+$ TDM network. In particular, we remove the lateral modules and use convolutional and upsampling operations from the top-down modules $\mathbf{T}$ to train `deeper' variants of VGG16 as baseline. To control for the extra parameters from lateral modules, we also increase the parameters in the convolutional layers. Note that for this baseline, we follow training methodology and design decisions used for training TDM networks. 

As shown in Table~\ref{tab:ablation}(left), even though using more depth increases the performance slightly, the performance boost due to lateral modules is much higher. This highlights the importance of dividing the capacity of TDM network amongst lateral and top-down modules.

\begin{table}[t]
\centering
\caption{(left) \textbf{Importance of lateral modules}: We use operations from the top-down module to increase depth of the VGG16 network; $\sim\mathbf{T}_{i,j}$ represents modified top-down module to account for more parameters. (right) \textbf{Impact of Pre-training}.\vspace{0.02in}}
\renewcommand{\arraystretch}{1.2}
\renewcommand{\tabcolsep}{1.2mm}
{\small
\begin{tabular}{L{0.7cm} C{1.2cm} C{1.4cm}@{}}
\toprule
{\footnotesize $i,j$} & {\footnotesize$\sim\mathbf{T}_{i,j}$} & {\footnotesize$\left(\mathbf{T}_{i,j}, \mathbf{L}_j\right)$} \\
\midrule
$5,4$ & 24.8 & 26.2 \\
$4,3$ & 25.1 & 28.8 \\
$3,2$ & 26.5 & \hl{29.9} \\ 
$2,1$ & 21.4 & 29.6 \\
\Xhline{1pt}
\end{tabular}
}
\quad
{\small
\begin{tabular}{L{0.8cm} C{1cm} C{1.5cm}@{}}
\toprule
& \multicolumn{2}{l}{\textbf{Pre-trained on:}} \\
& COCO & ImageNet \\
\midrule
$\mathbf{T}_{4,3}$ & 34.4 & 34.0 \\
$\mathbf{T}_{3,2}$ & 35.3 & 34.1 \\
\Xhline{1pt}
\\
\end{tabular}
}
\vspace{-0.05in}
\label{tab:ablation}
\end{table}

\subsection{No top-down modules}\label{sec:notop}
Next we want to study the importance of the top-down path introduced by our TDM network. We believe that this path is responsible for transmitting contextual features and for selection of relevant finer details. Removing the top-down path exposes the `skip'-connections from bottom-up features, which can be used for object detection. We follow the strategy from~\cite{bell2015inside}, where they ROI-pool features from different layers, L2-normalize and concatenate these features and finally scale them back to the original \conv\verb|5_3| magnitude and dimension.

We tried many variants of the Skip-pooling baseline, and report the best results in Table~\ref{tab:coco_minival} (Skip-pool). We see that performance for small objects (AP$^\text{S}$) increases slightly, but overall the AP does not change much. This highlights the importance of using high-level contextual features in the top-down path for the selection of low-level features.

\subsection{Impact of Pre-training}\label{sec:pretraining} 
Finally, we study the impact of using a model pre-trained on the detection task to initialize our base networks and ResNet101/InceptionResNetv2's RPN and RCN networks \vs using only an image classification~\cite{imagenet} pre-trained model. In Table~\ref{tab:ablation}(right), we see that initialization does not impact the performance by a huge margin. However, pre-training on the detection task is consistently better than using the classification initialization.

\section{Conclusion}
This paper introduces the Top-Down Modulation (TDM) network, which leverages top-down contextual features and lateral connections to bottom-up features for object detection. The TDM network uses top-down context to select low-level finer details, and learns to integrate them together. Through extensive experiments on the challenging COCO dataset, we demonstrate the effectiveness and importance of features from the TDM network. We show empirically that the proposed representation benefits all objects, big and small, and is helpful for accurate localization.

Even though we focused on the object detection, we believe these top-down modulated features will be helpful in a wide variety of computer vision tasks.

{\small
\bibliographystyle{ieee}
\bibliography{egbib}
}

\end{document}